\def\year{2019}\relax
\documentclass[letterpaper]{article} 
\usepackage{aaai19}  
\usepackage{times}  
\usepackage{helvet}  
\usepackage{courier}  
\usepackage{url}  
\usepackage{graphicx}  
\usepackage{booktabs} 
\usepackage{subfigure}
\usepackage{color}
\usepackage{multirow}
\usepackage{mathrsfs}
\usepackage{enumitem}
\usepackage{makecell}
\usepackage{tabulary}
\usepackage{pifont}
\usepackage{amsfonts}
\begin{document}
%
\title{Temporal Deformable Convolutional Encoder-Decoder Networks\\for Video Captioning\protect \thanks{This work was performed at JD AI Research.}}

\author{Jingwen Chen \textsuperscript{1},
Yingwei Pan \textsuperscript{2},
Yehao Li \textsuperscript{1},
Ting Yao \textsuperscript{2},
Hongyang Chao \textsuperscript{1,3},
Tao Mei \textsuperscript{2}\\
\textsuperscript{1}Sun Yat-sen University, Guangzhou, China\\
\textsuperscript{2}JD AI Research, Beijing, China\\
\textsuperscript{3}The Key Laboratory of Machine Intelligence and Advanced Computing (Sun Yat-sen University),\\Ministry of Education, Guangzhou, China\\
\{chenjingwen.sysu, panyw.ustc, yehaoli.sysu, tingyao.ustc\}@gmail.com,
isschhy@mail.sysu.edu.cn,
tmei@live.com}

\maketitle
\begin{abstract}
It is well believed that video captioning is a fundamental but challenging task in both computer vision and artificial intelligence fields. The prevalent approach is to map an input video to a variable-length output sentence in a sequence to sequence manner via Recurrent Neural Network (RNN). Nevertheless, the training of RNN still suffers to some degree from vanishing/exploding gradient problem, making the optimization difficult. Moreover, the inherently recurrent dependency in RNN prevents parallelization within a sequence during training and therefore limits the computations. In this paper, we present a novel design --- Temporal Deformable Convolutional Encoder-Decoder Networks (dubbed as TDConvED) that fully employ convolutions in both encoder and decoder networks for video captioning. Technically, we exploit convolutional block structures that compute intermediate states of a fixed number of inputs and stack several blocks to capture long-term relationships. The structure in encoder is further equipped with temporal deformable convolution to enable free-form deformation of temporal sampling. Our model also capitalizes on temporal attention mechanism for sentence generation. Extensive experiments are conducted on both MSVD and MSR-VTT video captioning datasets, and superior results are reported when comparing to conventional RNN-based encoder-decoder techniques. More remarkably, TDConvED increases CIDEr-D performance from 58.8\% to~67.2\% on MSVD.
\end{abstract}

\section{Introduction}
\begin{figure}[!tb]
\centering
\includegraphics[width=0.40\textwidth]{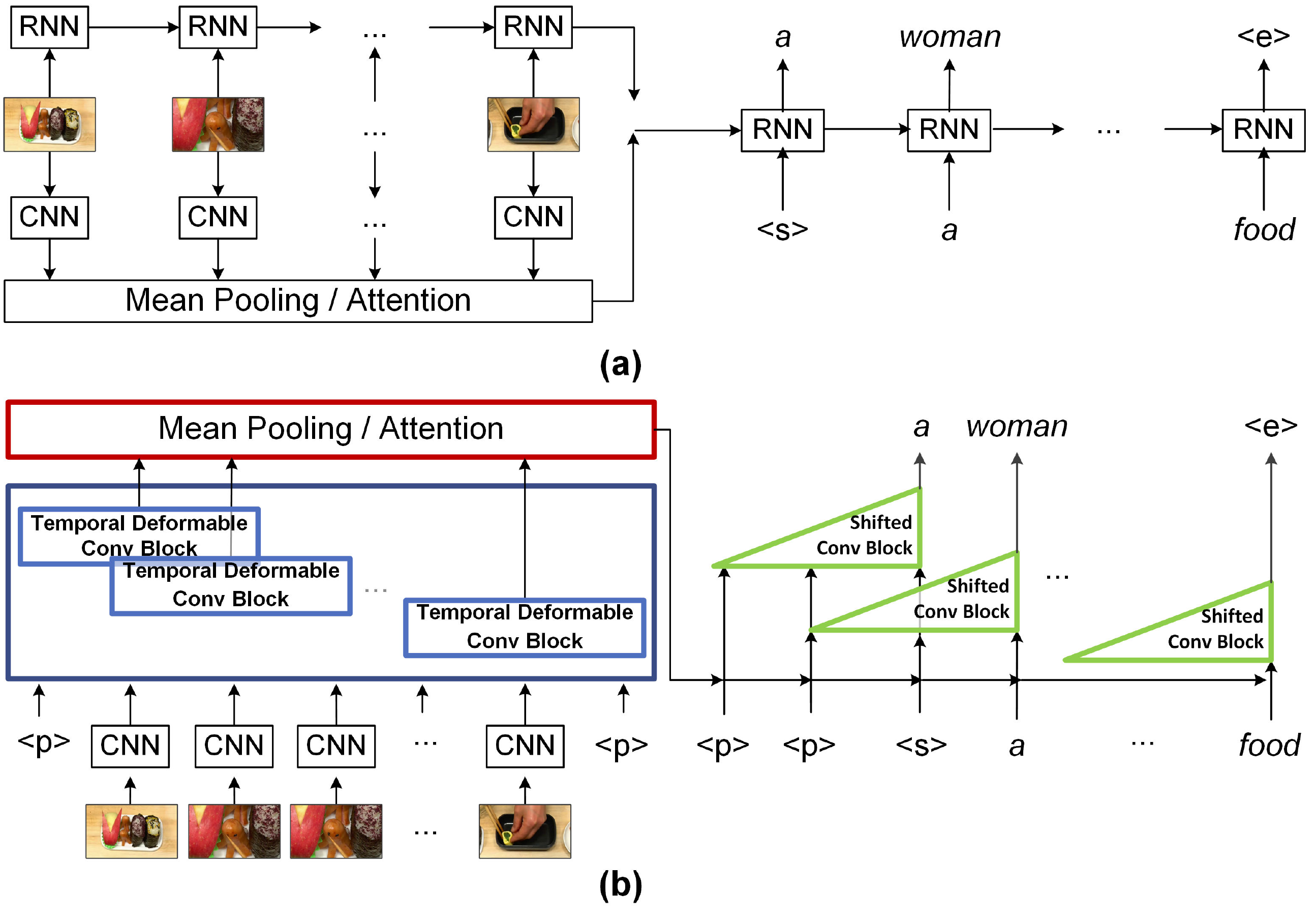}
\caption{An illustration of (a) CNN or RNN plus RNN architecture and (b) our Temporal Deformable Convolutional Encoder-Decoder Networks, for video captioning.}
\label{fig:intro}
\end{figure}

The recent advances in deep neural networks have convincingly demonstrated high capability in learning vision models particularly for recognition. The achievements make a further step towards the ultimate goal of video understanding, which is to automatically describe video content with a complete and natural sentence or referred to as video captioning problem. The typical solutions \cite{Pan:CVPR16,Venugopalan:ICCV15,Venugopalan:NAACL15,Yao:ICCV15,Yu:CVPR16} of video captioning are inspired by machine translation and equivalent to translating a video to a text. As illustrated in Figure \ref{fig:intro} (a), the basic idea is to perform sequence to sequence learning, where a Convolutional Neural Network (CNN) or Recurrent Neural Network is usually exploited to encode a video and a decoder of RNN is utilized to generate the sentence, one word at each time step till the end. Regardless of different versions of CNN or RNN plus RNN video captioning framework, a common practice is the use of sequential RNN, which may result in two limitations. First, RNN powered by Long Short-Term Memory (LSTM) alleviates the issue of vanishing or exploding gradients to some extent, nevertheless the problem still exists. Moreover, the utilization of RNN on a long path between the start and end of a sequence will easily forget the long-range information. Second, the training process of RNN is inherently sequential due to the recurrent relation in time and thus prevents parallel computations in a sequence.

We propose to mitigate the aforementioned two issues via convolutions in an encoder-decoder architecture. Compared to the chain structure in RNN, feed-forward convolutions produce representations of fixed-length inputs and could easily capture long-term dependencies by stacking several layers on the top of each other. Unlike RNN in which the inputs are fed through a variable number of non-linearities and the number depends on the input length, the operation of convolution fixes the number of non-linearities and thus eases the optimization. Furthermore, feed-forward convolutions are without any recurrent functions and do not rely on the computation of the previous time step. Therefore, computations over all the input elements could be fully parallelized during training.

By consolidating the idea of leveraging convolutions in sequence-to-sequence learning for video captioning, we present a novel Temporal Deformable Convolutional Encoder-Decoder Networks (TDConvED) architecture, as conceptually shown in Figure \ref{fig:intro} (b). Given a video, a standard CNN is utilized to extract visual features of sampled frames or clips. Then, the features of frames/clips are fed into convolutional block to produce intermediate states of each frame/clip with contexts. The context size is enlarged by stacking several convolutional blocks. Please also note that we endow the convolutional block in encoder with more power of devising temporal deformable convolutions by capturing dynamics in temporal extents of actions or scenes. Once all sampled frames/clips are encoded, the video-level representations are computed by mean pooling over all the intermediate states. Similarly, we employ feed-forward convolutional blocks that operate on the concatenation of video-level representations and the embeddings of past words to generate subsequent word, till the end token is predicted. In addition, sentence generation is further enhanced by additionally incorporating temporally attended video features. Since there are no recurrent connections and all ground-truth words are available at any time step, our TDConvED model could be trained in parallel.

Our main contribution is the proposal of the use of convolutions for sequence-to-sequence learning and eventually boosting video captioning. This also leads to the elegant view of how to design feed-forward convolutional structure in an encoder-decoder video captioning framework, and how to support parallelization over every element within a sequence, which are problems not yet fully~understood.

\section{Related Work}

\textbf{Video Captioning.}
The dominant direction in modern video captioning is sequence learning approaches \cite{Pan:CVPR16,Pan:CVPR17,chen2017video,Venugopalan:ICCV15,Venugopalan:NAACL15,Yao:ICCV15,Yu:CVPR16,li2018jointly} which utilize RNN-based architecture to generate novel sentences with flexible syntactical structures. For instance, Venugopalan \emph{et al.} present a LSTM-based model to generate video descriptions with the mean pooled representation over all frames in \cite{Venugopalan:NAACL15}. The framework is then extended by inputting both frames and optical flow images into an encoder-decoder LSTM in \cite{Venugopalan:ICCV15}. Compared to mean pooling, Yao \emph{et al.} propose to utilize the temporal attention mechanism to exploit temporal structure for video captioning \cite{Yao:ICCV15}. Later in \cite{Zhao18:tubefeatures}, Zhao \emph{et al.} design an object-aware tube feature for video captioning to enable attention on salient objects. Most recently, Hao \emph{et al.} develop several deep fusion strategies to effectively integrate both visual and audio cues into sentence generation in \cite{Hao:aaai18}.

\begin{figure*}[!tb]
\centering
\includegraphics[width=0.80\textwidth]{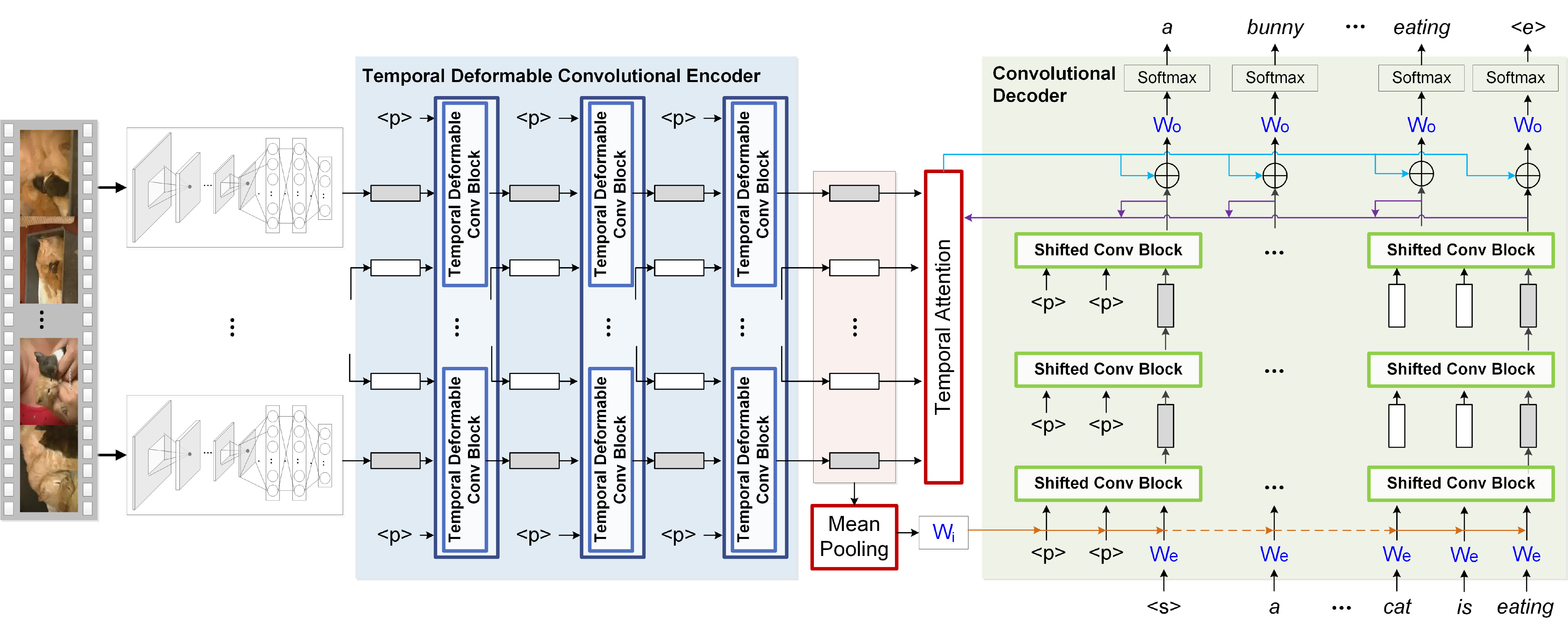}
\caption{The overall architecture of our TDConvED that fully employs convolutions in both encoder and decoder networks for video captioning. A standard CNN is firstly utilized to extract visual features of sampled frames/clips. Then, the features of frames/clips are fed into a temporal deformable convolutional encoder with stacked temporal deformable convolutional blocks to produce intermediate states of each frame/clip with contexts. The video-level representations are computed by mean pooling over all the intermediate states. After that, a convolutional decoder with stacked shifted convolutional blocks operates on the concatenation of video-level representations and the embeddings of past words to generate the next word. In addition, a temporal attention mechanism tailored to such convolutional encoder-decoder structure is incorporated to boost video captioning.}
\label{fig:overview}
\end{figure*}

\textbf{Sequence Learning with CNN.}
The most typical paradigm in sequence learning is RNN-based encoder-decoder structure which mainly capitalizes on RNN to model the probability of decoding an output given the previous outputs and all the inputs. Although the remarkable results have been observed on a number of sequential tasks (e.g., image/video captioning, machine translation), the inherent recursive characteristic inevitably limits the parallelization abilities and even raises vanishing/exploding gradient problems in the training stage. To tackle these barriers, there is an emerging trend of leveraging Convolutional Neural Network (CNN) for sequence learning in language modeling for NLP tasks \cite{Meng:ACL15,Bradbury:QRNN16,Kalchbrenner:ByteNet16,Gehring:ICML17}. A convolutional encoder with a gated architecture has been designed in \cite{Meng:ACL15}, which could pinpoint the parts of a source sentence that are relevant to the target word for machine translation. Recently, the first fully convolutional model for sequence learning is proposed in \cite{Gehring:ICML17} to design both encoder and decoder in the form of convolutions with CNN, which even outperforms strong recurrent models on machine translation task.

\textbf{Summary.}
In short, our approach in this paper belongs to sequence learning method for video captioning. Unlike most of the aforementioned sequence learning models which mainly focus on generating sentence by solely depending on RNN-based decoder, our work contributes by exploiting a fully convolutional sequence learning architecture that relies on CNN-based encoder and decoder for video captioning. Moreover, we additionally explore the temporal deformable convolutions and temporal attention mechanism to extend and utilize temporal dynamics across frames/clips, and eventually enhance the quality of generated sentence.

\section{Temporal Deformable Convolutional~Encoder-Decoder~Networks}
Our Temporal Deformable Convolutional Encoder-Decoder Networks (TDConvED) architecture is devised to generate video descriptions by fully capitalizing on convolutional encoder and decoder. Specifically, a convolutional encoder is firstly leveraged to encode each frame/clip with the contextual information among multiple sampled frames/clips via temporal deformable convolution. The idea behind temporal deformable convolution is to augment the temporal samplings with additional offsets learnt from the inherent temporal dynamics of scenes/actions within frame/clip sequence. Next, conditioned on the video-level representation induced by the temporal deformable convolutional encoder (e.g., performing mean pooling over all encoded features of frames/clips), we employ a convolutional decoder to predict each output word by additionally exploiting the contextual relations across the previous output words. Furthermore, a temporal attention mechanism tailored to temporal deformable convolutional encoder-decoder structure is designed to further boost video captioning. An overview of TDConvED is shown in Figure~\ref{fig:overview}.

\subsection{Problem Formulation}
Suppose a video $V$ with $N_v$ sampled frames/clips (uniform sampling) to be described by a textual sentence $S$, where $S = (w_1, w_2, \dots, w_{N_s})$ is the word sequence consisting of $N_s$ words. Let ${\bf{v}} = (v_1, v_2, \dots, v_{N_v})$ denote the temporal sequence of frame/clip representations of the video $V$, where $v_i \in \mathbb{R}^{D_v}$ represents the $D_v$-dimensional feature of $i$-th frame/clip extracted by 2D/3D CNN.

Inspired by the recent successes of sequence to sequence solutions leveraged in image/video captioning \cite{Venugopalan:ICCV15,yao2017novel,yao2017boosting,yao2018exploring}, we aim to formulate our video captioning model in an encoder-decoder scheme. Instead of relying on RNN to compute intermediate encoder/decoder states, we novelly utilize a fully convolutional encoder-decoder architecture for sequence to sequence modeling in video captioning. Formally, a temporal deformable convolutional encoder is utilized to encode the input frame/clip sequence ${\bf{v}} = (v_1, v_2, \dots, v_{N_v})$ into a set of context vectors ${\bf{z}} = (z_1, z_2, \dots, z_{N_v})$, which are endowed with the contextual information among frames/clips sampled in a free-form temporal deformation along the input sequence. Next, we perform mean pooling over all the context vectors ${\bf{z}}$ to achieve the video-level representation, which is further injected into the convolutional decoder to generate target output sentence $S$ by capturing the long-term contextual relations among input video and previous output words. Thus, the video sentence generation problem we exploit is formulated by minimizing the following energy loss:
\begin{equation}\label{Eq:EqPF1}\small
E({\bf{v}}, S) = -\log {\Pr{(S|{\bf{v}})}},
\end{equation}
which is the negative $\log$ probability of the correct textual sentence given the temporal sequence of input video. By applying chain rule to model the joint probability over the sequential words, the $\log$ probability of the sentence is measured by the sum of the $\log$ probabilities over each word, which can be written as
\begin{equation}\label{Eq:Eq2}\small
\log {\Pr{(S|{\bf{v}})}} =  \sum\limits_{t = 1}^{{N_s}} {\log \Pr\left( {\left. {{{{w}}_t}} \right|{\bf{v}}, {{{w}}_0}, \ldots ,{{{w}}_{t - 1}}} \right)}.
\end{equation}
In contrast to the conventional RNN-based encoder-decoder architecture that models the above parametric distribution $\Pr\left( {\left. {{{{w}}_t}} \right|{\bf{v}}, {{{w}}_0}, \ldots ,{{{w}}_{t - 1}}} \right)$ with RNN, we employ convolutions to calculate it in a simple feed-forward manner at decoding stage. Since no recurrent dependency exists in the convolutional decoder and all grough-truth words are available at training, the joint probability ${\Pr{(S|{\bf{v}})}}$ can be calculated in parallel for all words. Derived from the idea of attention mechanism in sequence to sequence modeling \cite{Yao:ICCV15}, we also capitalize on temporal attention mechanism to focus on the frames/clips that are highly relevant to the output word for enriching the video-level representation. Such representation will be additionally injected into convolutional decoder to enhance video captioning.

\subsection{Temporal Deformable Convolutional Encoder}
\begin{figure}
\centering
\includegraphics[width=0.36\textwidth]{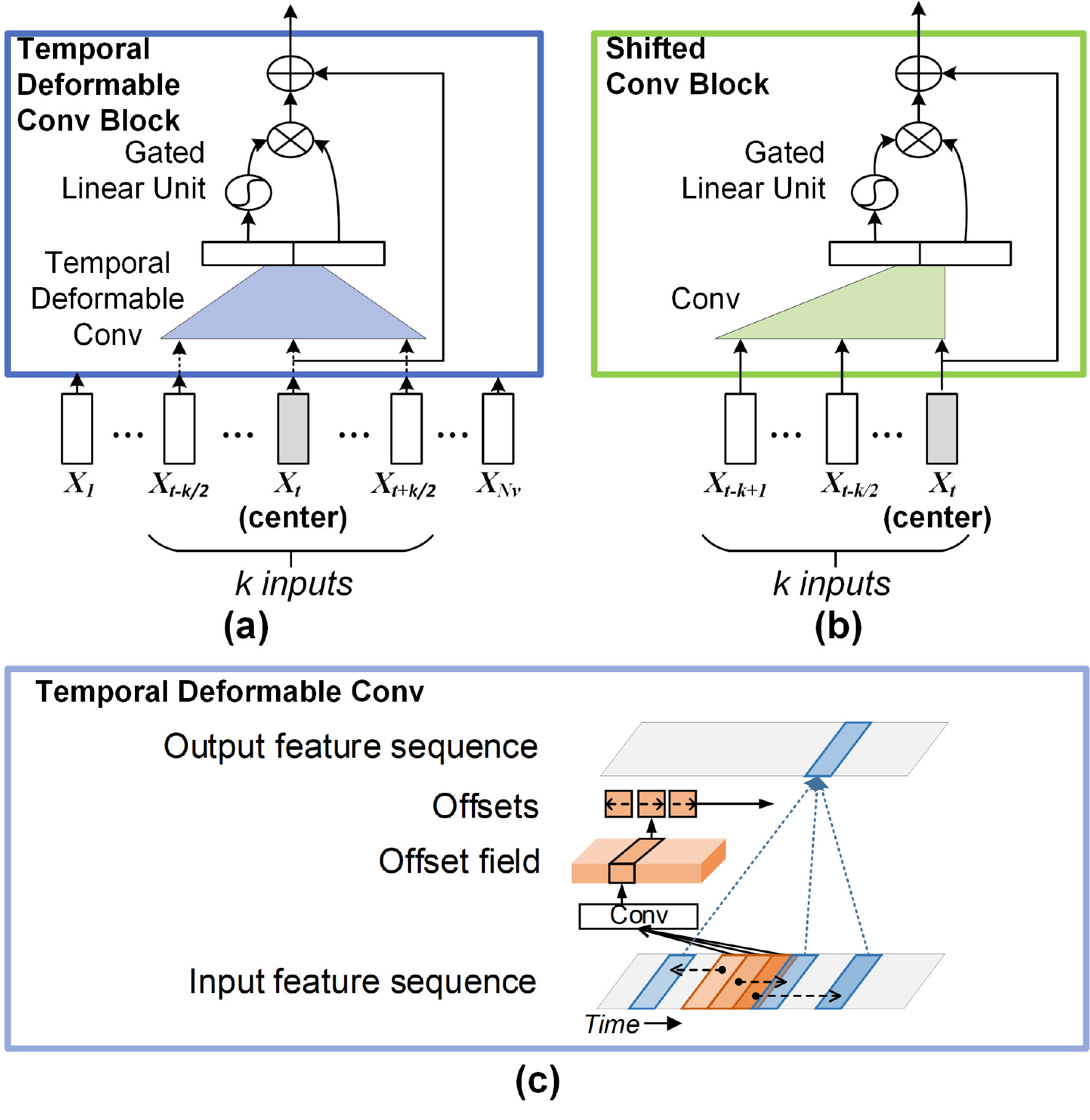}
\caption{An illustration of (a) the temporal deformable conv block in encoder, (b) the shifted conv block in decoder, and (c) the temporal deformable convolution.}
\label{fig:conv}
\end{figure}

An encoder is a module that takes the source sequence (i.e., frame/clip sequence of video) as input and produces intermediate states to encode the semantic content. Here we devise a temporal deformable convolutional block in the encoder of our TDConvED which applies temporal deformable convolution over the input sequence to capture the context across frames/clips sampled in a free-form temporal deformation, as depicted in Figure \ref{fig:conv} (a). Such design of temporal deformable convolution improves conventional temporal convolution by capturing temporal dynamics on the natural basis of actions/scenes within the video. Meanwhile, the feed-forward convolutional structure in encoder enables the parallelization within the input sequence and allows fast computation. Furthermore, to exploit long-term dependency among input sequence at encoding stage, multiple temporal deformable convolutional blocks are stacked in encoder to integrate the contextual information from a larger number of temporal samplings in the input sequence.

Formally, consider the $l$-th temporal deformable convolutional block in encoder with the corresponding output sequence denoted as ${\bf{p}}^l = (p^l_1, p^l_2, \dots, p^l_{N_v})$, where $p^l_i \in \mathbb{R}^{D_r}$ is the output of the temporal deformable convolution centered on the $i$-th frame/clip. Given the output sequence ${\bf{p}}^{l-1} = (p^{l-1}_1, p^{l-1}_2, \dots, p^{l-1}_{N_v})$ of the $(l-1)$-th block, each output intermediate state $p^l_i$ is achieved by feeding the subsequence of ${\bf{p}}^{l-1}$ into a temporal deformable convolution (kernel size: $k$) plus a non-linearity unit. Note that the temporal deformable convolution operates in a two-stage way, i.e., first measuring temporal offsets via an one-dimensional convolution for sampling frames/clips and then aggregating the features of sampled frames/clips, as depicted in Figure \ref{fig:conv} (c). More specifically, let $X = (p^{l-1}_{i + r_1}, p^{l-1}_{i + r_2}, \dots, p^{l-1}_{i + r_k})$ denote the subsequence of ${\bf{p}}^{l-1}$, where $r_n$ is the $n$-th element of $R$ and $R = \{- k / 2, \dots, 0, \dots, k / 2\}$. The one-dimensional convolution in the $l$-th temporal deformable convolutional block can be parameterized as the transformation matrix $W^l_f \in \mathbb{R}^{k \times kD_r}$ and the bias $b^l_f \in \mathbb{R}^{k}$, which takes the concatenation of $k$ elements in $X$ as input and produces a set of offsets $\Delta r^i = \{\Delta r^i_n\}^{k}_{n=1}\in \mathbb{R}^{k}$:
\begin{equation}\small
\Delta r^i = W^l_f [p^{l-1}_{i + r_1}, p^{l-1}_{i + r_2}, \dots, p^{l-1}_{i + r_k}] + b^l_f,
\label{eq:conv1}
\end{equation}
where the $n$-th element $\Delta r^i_n$  in $\Delta r^i$ denotes the measured temporal offset for the $n$-th sample in the subsequence $X$. Next, we achieve the output of temporal deformable convolution by augmenting samples with temporal offsets via another one-dimensional convolution:
\begin{equation}\small
o^l_i = W^l_d [p^{l-1}_{i + r_1 + \Delta r^i_1}, p^{l-1}_{i + r_2 + \Delta r^i_2}, \dots,  p^{l-1}_{i + r_k + \Delta r^i_k}] + b^l_d,
\label{eq:defconv}
\end{equation}
where $W^l_d \in \mathbb{R}^{2D_r \times kD_r}$ is the transformation matrix in the one-dimensional convolution and $b^l_d \in \mathbb{R}^{2D_r}$ is the bias. As the temporal offset $\Delta r^i_n$ is typically fractional, $p^{l-1}_{i + r_n + \Delta r^i_n}$ in Eq.(\ref{eq:defconv}) can be calculated via temporal linear interpolation:
\begin{equation}\small
p^{l-1}_{i + r_n + \Delta r^i_n} = \sum\limits_{s} {B(s, i + r_n + \Delta r^i_n) p^{l-1}_s},
\end{equation}
where $i + r_n + \Delta r^i_n$ denotes an arbitrary position, $s$ enumerates all integral positions in the input sequence ${\bf{p}}^{l-1}$, and $B(a, b) = max(0, 1 - |a - b|)$.

In addition, we adopt a gated linear unit (GLU) \cite{Dauphin:ICML17} as the non-linearity unit to ease gradient propagation. Thus given the output of temporal deformable convolution $o^l_i\in \mathbb{R}^{2D_r}$ which has twice the dimension of the input element, a GLU is applied over $o^l_i=[A,B]$ through a simple gating mechanism:
\begin{equation}\small
g(o^l_i) = A \otimes \sigma(B) ,
\label{eq:glu}
\end{equation}
where $A, B \in \mathbb{R}^{D_r}$ and $\otimes$ is the point-wise multiplication. $\sigma(B)$ represents a gate unit that controls which elements of $A$ are more relevant to the current context. Moreover, the residual connections \cite{He:CVPR16} from input of the temporal deformable convolution block to the output of the block are added to enable deeper networks. Hence, the final output of the $l$-th temporal deformable convolutional block is measured as
\begin{equation}\small
p^l_i = g(o^l_i) + p^{l-1}_i.
\label{eq:conv2}
\end{equation}

To ensure the match between the length of output sequence of temporal deformable convolutional block and the input length, both the left and right sides of the input are padded with $k/2$ zero vectors. By stacking several temporal deformable convolutional blocks over the input frame/clip sequence, we obtain the final sequence of context vectors ${\bf{z}} = (z_1, z_2, \dots, z_{N_v})$, where $z_i \in \mathbb{R}^{D_r}$ represents the contextually encoded feature of the $i$-th frame/clip.

\subsection{Convolutional Decoder}
The most typical way to implement the decoder for sequence learning is to adopt sequential RNN to produce the output sequence conditioned on the representations of the input sequence induced by the encoder. However, the inherent dependency in RNN inevitably limits the parallelization abilities of RNN-based decoder and even raises vanishing/exploding gradient problems for training. To alleviate the challenges in decoder of sequence learning, we devise a fully convolutional decoder by stacking several shifted convolutional blocks to capture long-term context across the encoded context vectors of input video and output words. Here, the center position of the convolution in shifted convolutional block is shifted with respect to the center of normal convolutions, as shown in Figure \ref{fig:conv} (b).

In particular, given the encoded sequence of context vectors ${\bf{z}} = (z_1, z_2, \dots, z_{N_v})$ for input video, we first perform mean pooling over all context vectors and achieve the video-level representation $\tilde{\bf{z}}=\frac{1}{N_v}\sum\limits_{i = 1}^{N_v} {z_i}$. Since no memory cells or autoregressive path exists in the convolutional decoder, the video-level representation $\tilde{\bf{z}}$ is transformed with a linear mapping $W_i$ and concatenated with the embeddings of the input word at each time step, which will be set as the input of the first shifted convolutional block of the decoder. The transformation for addition of word embeddings and position embeddings is denoted as $W_e$. Consider the $l$-th shifted convolutional block in decoder with the corresponding output sequence ${\bf{q}}^l = (q^l_0, q^l_1, \dots, q^l_{N_s})$, where $q^l_t \in \mathbb{R}^{D_f}$ denotes the output of the convolution at time step $t$. More specifically, by taking the output sequence of the $(l-1)$-th shifted convolutional block as input, every subsequence of ${\bf{q}}^{l-1}$ is fed into an one-dimensional convolution (kernel size: $k$) plus a GLU and residual connections, resulting in the output intermediate state $q^l_t$ at time step $t$. Different from the basic convolution, the corresponding subsequence of ${\bf{q}}^{l-1}$ is $X_q = (q^{l-1}_{t - k + 1}, q^{l-1}_{t - k + 2}, \dots, q^{l-1}_t)$ since no future information is available to decoder at decoding stage. Hence, the operation in each shifted convolutional block is given by
\begin{equation}\small
q^l_t = g(W^q_l[q^{l-1}_{t - k + 1}, q^{l-1}_{t - k + 2}, \dots, q^{l-1}_t] + b^q_l) + q^{l-1}_t,
\end{equation}
where $W^q_l \in \mathbb{R}^{2D_f \times kD_f}$ and $b^q_l \in \mathbb{R}^{2D_f}$ denote the transformation matrix and bias of the convolutional kernel in the $l$-th shifted convolutional block, respectively. Note that for each shifted convolutional block, only the left side of the input sequence is padded with $k - 1$ zero vectors for matching the length between input and output sequence. Accordingly, by leveraging the convolutional decoder with stacked shifted convolutional blocks, we obtain the final output sequence ${\bf{h}} = (h_1, h_2, \dots, h_{N_s})$ and each element $h_t \in \mathbb{R}^{D_f}$ denotes the output intermediate state at time step $t$.

\subsection{Temporal Attention Mechanism}\label{sec:TA}
In many cases, the output word at decoding stage only relates to some frames/clips of the input video. As a result, utilizing one encoder to compress the whole video into a global feature (e.g., the aforementioned mean pooled video-level representation $\tilde{\bf{z}}$) would collapse the natural temporal structure within video and bring in the noises from irrelevant frames/clips, resulting in sub-optimal results. To dynamically pinpoint the frames/clips that are highly relevant to the output word and further incorporate the contributions of different frames/clips into producing video-level representation for predicting target word, a temporal attention mechanism is employed over the encoded sequence of context vectors ${\bf{z}} = (z_1, z_2, \dots, z_{N_v})$  to boost video captioning. Specifically, at each time step $t$, the attention mechanism first generates a normalized attention distribution over all the encoded context vectors ${\bf{z}}$ depending on the current output intermediate state $h_t$ of the convolutional decoder as
\begin{equation}\small
a^t_i={W}_a\left[\tanh\left({W}_z{{z}_i} + {W}_h{h}_t+ b_a \right)\right],\lambda^t=softmax\left({a}^t\right),
\end{equation}
where $a^t_{i}$ is the $i$-th element of $a^t$, $W_a\in {{\mathbb{R}}^{{1}\times {D_a}}}$, $W_z\in {{\mathbb{R}}^{{D_a}\times {D_r}}}$ and $W_h\in {{\mathbb{R}}^{{D_a}\times {D_f}}}$ are transformation matrices. $\lambda^t \in\mathbb R^{N_v}$ denotes the normalized attention distribution and its $i$-th element $\lambda^t_i$ is the attention probability of context vector $z_i$ of the $i$-th frame/clip. Based on the attention distribution, we calculate the attended video-level feature $\hat {\bf{z}}_t$ by aggregating all the context vectors weighted by attention:
\begin{equation}\small
\hat {\bf{z}}_t  = \sum\limits_{i = 1}^{N_v} {\lambda^t_i z_i}.
\end{equation}
The attended video-level feature $\hat {\bf{z}}_t$ is further transformed with a linear mapping and combined with the output intermediate state $h_t$ of convolutional decoder, which is employed to predict the next word through a softmax layer.

\subsection{Inference}
Similar to RNN based model, the inference of TDConvED is performed sequentially by generating one word at each time step. At the beginning of inference, the start sign word $<s>$ and the encoded sequence ${\bf{z}} = (z_1, z_2, \dots, z_{N_v})$ of input video are jointly fed into the convolutional decoder to predict the output word distribution. The word with the maximum probability is popped out as the input word to the convolutional decoder at the next time step. This process continues until the end sign word $<e>$ is emitted or the pre-defined maximum sentence length is reached.

\section{Experiments}
We evaluate and compare our TDConvED with conventional RNN-based techniques by conducting video captioning task on two benchmarks, i.e., Microsoft Research Video Description Corpus (MSVD) \cite{Chen:ACL11} and Microsoft Research Video to Text (MSR-VTT) \cite{Xu:CVPR16}.

\subsection{Datasets and Experimental Settings}
\textbf{MSVD.} The MSVD dataset is the most popular video captioning benchmark and contains 1970 short videos clips from YouTube. Following the standard settings in previous works \cite{Pan:CVPR16,Yao:ICCV15}, we take 1200 videos for training, 100 for validation and 670 for testing.

\textbf{MSR-VTT.}
MSR-VTT is a recently released large-scale benchmark for video captioning and consists of 10,000 web video clips from 20 well-defined categories. Following the official split, we utilize 6513, 497, and 2990 video clips for training, validation and testing, respectively.

\textbf{Features and Parameter Settings.}
We uniformly sample 25 frames/clips for each video and each word in the sentence is represented as ``one-hot" vector (binary index vector in a vocabulary). For visual representations, we take the output of 4,096-way fc7 layer from VGG or 2,048-way pool5 from ResNet \cite{He:CVPR16} pre-trained on ImageNet \cite{Russakovsky:IJCV15}, and 4,096-way fc6 layer from C3D \cite{Tran:ICCV15} pre-trained on Sports-1M \cite{Karpathy:CVPR14} as frame/clip representations. In training phase, we add a start token $<s>$ to indicate the starting of sentence, an end token $<e>$ to denote the end of each sentence, and a padding token $<p>$ to keep the length of all textual sequences the same. For sentence generation in inference stage, we adopt the beam search and set the beam size as 5. The kernel size $k$ of convolutions in encoder and decoder is set as 3. The convolutional encoder/decoder in our TDConvED consists of 2 stacked temporal deformable/shifted convolutional blocks. Both the dimensions of intermediate states in encoder and decoder, i.e., $D_r$ and $D_f$, are set as 512. The dimension of the hidden layer for measuring attention distribution $D_a$ is set as 512. The whole model is trained by Adam \cite{kingma2014adam} optimizer. We set the initial learning rate as $10 ^ {-3}$ and the mini-batch size as 64. The maximum training iteration is set as 30~epoches.

\begin{table}[tb]\small
\centering
\caption{Performances of TDConvED and other RNN-based approaches on MSVD, where B@4, M and C are short for BLEU@4, METEOR and CIDEr-D. All values are reported as percentage (\%). The short name in the brackets indicates the features, where G, V, C, O, R and M denotes GoogleNet, VGGNet, C3D, Optical flow, ResNet and motion feature learnt by 3D CNN on hand-crafted descriptors, respectively.}
\begin{tabular}{lccc}
\Xhline{2\arrayrulewidth}
Model &M & C & B@4 \\
\hline\hline
MP-LSTM (V)  &29.2 & 53.3 & 37.0 \\
MP-LSTM (V+C) &29.7 & 55.1 & 39.4\\
MP-LSTM (R) & 32.5 & 71.0 & 50.4 \\\hline
S2VT (V+O) & 29.8 & - & - \\
S2VT (V+C) & 30.0 & 58.8 & 42.1 \\\hline
TA (G+M) & 29.6 & 51.7 & 41.9 \\
TA (V+C) & 29.9 & 56.1 & 40.9 \\
TA (R) & 33.3 & 72.0 & 51.3 \\ \hline
LSTM-E (V+C)  &31.0 & - & 45.3 \\
GRU-RCN (G)  & 31.6 & 68.0 & 43.3 \\
BAE (R+C)  & 32.4 & 63.5 & 42.5 \\
p-RNN (V+C) & 32.6 & 65.8 & 49.9 \\
MAMRNN (G)  & 32.9 & - & - \\
PickNet-VL (R)  & 33.1 & 76.0 & 46.1 \\
M$^3$ (G+C)  &33.3 & - & 52.8 \\
TF (R)  & 33.4 & - & - \\
\hline
TDConvED$_1$ (V) &29.4 &53.6 &41.0 \\
TDConvED$_2$ (V) &30.6 &58.1 &44.1 \\
TDConvED (V) &30.8 &58.2 &45.2 \\
TDConvED$_1$ (V+C) &31.7 &61.6 &45.9 \\
TDConvED$_2$ (V+C) &32.5 &65.3 &47.8 \\
TDConvED (V+C) &32.7 &67.2 &49.8 \\
TDConvED$_1$ (R) &32.9 &72.3 &48.3 \\
TDConvED$_2$ (R) &33.3 &74.6 &51.2 \\
TDConvED (R) &33.8 &76.4 &53.3 \\
\Xhline{2\arrayrulewidth}
\end{tabular}
\label{tab:msvd}
\vspace{-0.1in}
\end{table}

\textbf{Evaluation Metrics.}
We adopt three common metrics: BLEU@4 \cite{Papineni:ACL02}, METEOR \cite{Banerjee:ACL05}, and CIDEr-D \cite{Vedantam:CVPR15}. All the metrics are computed using the API \footnote{https://github.com/tylin/coco-caption} released by Microsoft COCO Evaluation Server \cite{Chen:coco15}.

\textbf{Compared Approaches.}
To empirically verify the merit of TDConvED, we compared our method with several conventional LSTM based models, including a CNN Encoder plus a LSTM Decoder model (\textbf{MP-LSTM}) \cite{Venugopalan:NAACL15}, LSTM based Encoder-Decoder model (\textbf{S2VT}) \cite{Venugopalan:ICCV15} and Temporal Attention based LSTM network (\textbf{TA}) \cite{Yao:ICCV15}. We also compared TDConvED with state-of-the-art methods: \textbf{GRU-RCN} \cite{Ballas:ICLR15}, \textbf{LSTM-E} \cite{Pan:CVPR16}, \textbf{p-RNN} \cite{Yu:CVPR16}, \textbf{MM} (only visual input) \cite{Ramanishka:ACMMM16}, \textbf{MA-LSTM} \cite{Xu:ACMMM17}, \textbf{BAE} \cite{Baraldi:CVPR17}, \textbf{hLSTMat} \cite{Song:IJCAI17}, \textbf{MAMRNN} \cite{Li17:MAMRNN}, \textbf{TF} \cite{Zhao18:tubefeatures}, \textbf{M$^3$} \cite{Wang18:M3}, \textbf{PickNet-VL} \cite{Chen18:PickNet} and \textbf{MCNN+MCF} \cite{Wu18:MCNN}. In addition to TDConvED, two different versions, named as TDConvED$_1$ and TDConvED$_2$, are also taken into comparisons. The former extends the basic CNN plus RNN model (MP-LSTM) by simply replacing the RNN-based decoder with our designed convolutional decoder, and the latter is more similar to TDConvED that only excludes temporal attention mechanism.

\begin{table}[!tb]\small
\centering
\caption{Performance of TDConvED and other RNN-based approaches on MSR-VTT, where B@4, M and C are short for BLEU@4, METEOR and CIDEr-D scores. All values are reported as percentage (\%). The short name in the brackets indicates the frame/clip features, where G, C, R and A denotes GoogleNet, C3D, ResNet and Audio feature.}
\begin{tabular}{lccc}
\Xhline{2\arrayrulewidth}
Model & M & C & B@4 \\
\hline\hline
MP-LSTM (R)  & 25.4 & 35.8 & 34.1 \\
MP-LSTM (G+C+A) & 25.6 & 38.1 & 35.7 \\\hline
S2VT (R)  & 25.8 & 36.7 & 34.4 \\
S2VT (G+C+A) & 26.0 & 39.1 & 36.0 \\\hline
TA (R)  & 24.9 & 34.5 & 33.2 \\
TA (G+C+A) & 25.1 & 36.7 & 34.8 \\\hline
LSTM-E (R) & 25.7 & 36.1 & 34.5 \\
LSTM-E (G+C+A) & 25.8 & 38.5 & 36.1 \\\hline
hLSTMat (R)  & 26.3 & - & 38.3 \\
MA-LSTM (G+C+A) & 26.5 & 41.0 & 36.5 \\
M$^3$ (V+C)  & 26.6 & - & 38.1 \\
MM (R+C)  & 27.0 & 41.8 & 38.3 \\
MCNN+MCF (R)  & 27.2 & 42.1 & 38.1 \\
PickNet-VL (R) & 27.2 & 42.1 & 38.9 \\
\hline
TDConvED$_1$ (R) & 26.8 & 40.7 & 37.1 \\
TDConvED$_2$ (R) & 27.2 & 41.9 & 39.0 \\
TDConvED (R) &27.5 &42.8 &39.5 \\
\Xhline{2\arrayrulewidth}
\end{tabular}
\label{tab:msrvtt}
\end{table}

\subsection{Experimental Results and Analysis}
\textbf{Quantitative Analysis.}
Table \ref{tab:msvd} shows the performances of different models on MSVD. It is worth noting that the reported performances of different approaches are often based on different frame/clip representations. For fair comparisons, our proposed models are evaluated on three commonly adopted frame/clip representations, i.e., output from VGG, the concatenation of features from VGG and C3D, and output from ResNet. Overall, the results across three evaluation metrics with the same input frame/clip representations consistently indicate that our proposed TDConvED achieves better results against three conventional RNN-based models including non-attention models (MP-LSTM and S2VT) and attention-based approach (TA). Specifically, by equipping decoder with convolutions, TDConvED$_1$ (V+C) makes the relative improvement over MP-LSTM (V+C) which adopts RNN-based decoder by 6.7\%, 11.8\% and 16.5\% in METEOR, CIDEr-D and BLEU@4, respectively. S2VT which exploits the context among the input frame/clip sequence through RNN-based encoder improves MP-LSTM, but the performances are still lower than our TDConvED$_2$. The result indicates the merit of leveraging temporal deformable convolutions in TDConvED$_2$ than RNN in encoder-decoder architecture for video captioning. Moreover, by additionally incorporating temporal attention mechanism into MP-LSTM model, TA leads to a performance boost. Similar in spirit, TDConvED improves TDConvED$_2$ by further taking temporal attention into account. There is a performance gap between TDConvED and TA. Though both runs involve the utilization of temporal attention, they are fundamentally different in the way that TA explores the attention distribution over original frame/clip features depending on the hidden state of RNN-based decoder, and TDConvED is by measuring attention distribution over the contextually encoded frame/clip features conditioned on the intermediate state of convolutional decoder. This again verifies the effectiveness of convolutional encoder-decoder architecture. In addition, compared to other state-of-the-art techniques (e.g., BAE, p-RNN, PickNet-VL, M$^3$, and TF), our TDConvED achieves better results in terms of all the three evaluation metrics. The results basically indicate that exploring convolutions in both encoder and decoder networks is a promising direction for video captioning.

The performance comparisons on MSR-VTT are summarized in Table \ref{tab:msrvtt}. Our TDConvED consistently outperforms other baselines. Similar to the observations on MSVD, TDConvED$_1$ exhibits better performance than the basic CNN plus RNN model, MP-LSTM, by replacing RNN with convolutions in decoder. Moreover, TDConvED$_2$ which additionally utilizing temporal deformable convolutional encoder to encode input frame/clip sequence performs better than TDConvED$_1$, and larger degree of improvement is attained when employing temporal attention mechanism in convolutional encoder-decoder structure by TDConvED.

\begin{figure}[!t]
\centering
\includegraphics[width=0.43\textwidth]{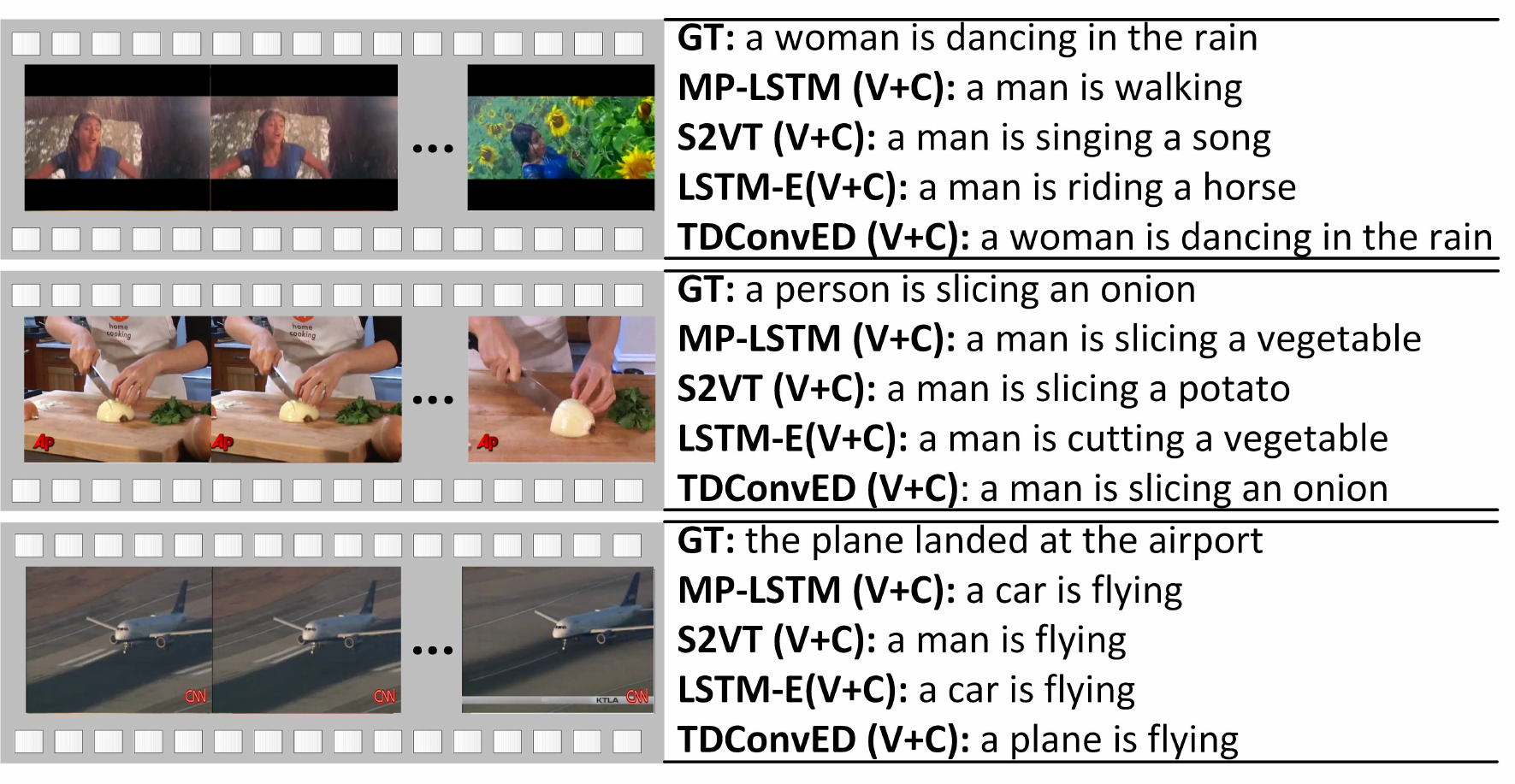}
\caption{Sentence generation results on MSVD dataset. The videos are represented by sampled frames and the output sentences are generated by 1) GT: ground truth sentences, 2) MP-LSTM, 3) S2VT, 4) LSTM-E and 5) our TDConvED.}
\label{fig:examples}
\end{figure}

\textbf{Qualitative Analysis.}
Figure \ref{fig:examples} shows a few sentence examples generated by different methods and human-annotated ground truth. It is easy to see that all of these automatic methods can generate somewhat relevant sentences, while TDConvED can predict more relevant keywords by exploiting long-range contextual relations via temporal deformable convolutional encoder-decoder structure and temporal attention mechanism. Compared to ``walking'' in the sentence generated by MP-LSTM, ``dancing'' in TDConvED is more precise to describe the first video. Similarly, the object ``onion'' presents the second video more exactly.

\textbf{Effect of the Number of Convolutional Blocks.}
In order to investigate the effect of the number of stacked convolutional blocks in our TDConvED, two additional experiments are conducted by varying the number of stacked convolutional blocks in encoder and decoder, respectively. As the encoder in TDConvED$_1$ is to perform mean pooling over the features of frames/clips and irrespective of convolutions, we first examine the impact of the convolutional block number in decoder of TDConvED$_1$. Table \ref{tab:layerdecoder} reports the comparison when the number ranges from 1 to 3. The best performances are achieved when the number of stacked convolutional blocks is 2. Next, by fixing the number of convolutional blocks in decoder as 2, Table \ref{tab:layerencoder} summarizes the performances of TDConvED$_2$ (V+C) when the convolutional block number in encoder varies from 1 to 3. The results show that the setting with 2 stacked temporal deformable convolutional blocks in encoder outperforms~others.

\begin{table}[!tb]\small
\centering
\caption{The effect of the number of stacked conv blocks (i.e., $\#$blocks) in (a) decoder and (b) encoder on MSVD.}
\subtable[Decoder in TDConvED$_1$]{
\begin{tabular}{ccc}
\Xhline{2\arrayrulewidth}
$\#$blocks & M & C \\
\hline\hline
1 & 31.6 & 57.9 \\
2 & 31.7 & 61.6 \\
3 & 31.4 & 59.8 \\
\Xhline{2\arrayrulewidth}
\end{tabular}
\label{tab:layerdecoder}
}
\qquad
\subtable[Encoder in TDConvED$_2$]{
\begin{tabular}{ccc}
\Xhline{2\arrayrulewidth}
$\#$blocks & M & C \\
\hline\hline
1 & 32.2 & 64.3 \\
2 & 32.5 & 65.3 \\
3 & 32.3 & 62.8 \\
\Xhline{2\arrayrulewidth}
\end{tabular}
\label{tab:layerencoder}
}
\label{tab:layeranalysis}
\end{table}

\begin{table}[!tb]\small
\centering
\caption{Comparison of training time between TDConvED and MP-LSTM on MSVD (Nvidia K40~GPU).}
\label{tab:time}
\begin{tabular}{lcccc}
\hline
Model & \#param & training time & B@4 & M \\
\hline
MP-LSTM (V) & 11.15 M & 2370 sec/epoch & 37.0 & 29.2 \\
TDConvED (V) & 16.79 M & 1070 sec/epoch & 45.2 & 30.8 \\
\hline
\end{tabular}
\end{table}

\textbf{Comparisons on Training Time.} Table \ref{tab:time} details the training time of our TDConvED and MP-LSTM. TDConvED is benefited from parallelization for training and the results clearly demonstrate the advantage of TDConvED at training speed, which is about 2.2$\times$ faster than MP-LSTM.

\section{Conclusions}
We have presented Temporal Deformable Convolutional Encoder-Decoder Networks (TDConvED) architecture, which explores feed-forward convolutions instead of RNN in sequence-to-sequence learning for video captioning. Particularly, we delve into the problem from the viewpoint of stacking several convolutional blocks to capture long-term contextual relationships for sequence learning. To verify our claim, we have built an encoder-decoder framework that encodes each frame/clip with contexts to intermediate states through multiple stacked convolutional blocks and then operates convolutional blocks on the concatenation of video features and the embeddings of past words to generate the next word. Moreover, through a new design of temporal deformable convolution, the encoder is further endowed with more power of capturing dynamics in temporal extents. As our encoder-decoder structure only involves feed-forward convolutions, the model allows parallel computation in a sequence at training. Extensive experiments conducted on both MSVD and MSR-VTT datasets validate our proposal and analysis. More remarkably, the improvements of our model are clear when comparing to RNN-based techniques.

\section{Acknowledgments}
This work is partially supported by NSF of China under Grant 61672548, U1611461, 61173081, and the Guangzhou Science and Technology Program, China, under Grant 201510010165.

\bibliography{sample-bibliography}

\begin{thebibliography}{}

\bibitem[\protect\citeauthoryear{Ballas \bgroup et al\mbox.\egroup
  }{2015}]{Ballas:ICLR15}
Ballas, N.; Yao, L.; Pal, C.; and Courville, A.~C.
\newblock 2015.
\newblock Delving deeper into convolutional networks for learning video
  representations.
\newblock {\em arXiv preprint arXiv:1511.06432}.

\bibitem[\protect\citeauthoryear{Banerjee and Lavie}{2005}]{Banerjee:ACL05}
Banerjee, S., and Lavie, A.
\newblock 2005.
\newblock {METEOR:} an automatic metric for {MT} evaluation with improved
  correlation with human judgments.
\newblock In {\em ACL}.

\bibitem[\protect\citeauthoryear{Baraldi, Grana, and
  Cucchiara}{2017}]{Baraldi:CVPR17}
Baraldi, L.; Grana, C.; and Cucchiara, R.
\newblock 2017.
\newblock Hierarchical boundary-aware neural encoder for video captioning.
\newblock In {\em CVPR}.

\bibitem[\protect\citeauthoryear{Bradbury \bgroup et al\mbox.\egroup
  }{2016}]{Bradbury:QRNN16}
Bradbury, J.; Merity, S.; Xiong, C.; and Socher, R.
\newblock 2016.
\newblock Quasi-recurrent neural networks.
\newblock {\em arXiv preprint arXiv:1611.01576}.

\bibitem[\protect\citeauthoryear{Chen and Dolan}{2011}]{Chen:ACL11}
Chen, D., and Dolan, W.~B.
\newblock 2011.
\newblock Collecting highly parallel data for paraphrase evaluation.
\newblock In {\em ACL}.

\bibitem[\protect\citeauthoryear{Chen \bgroup et al\mbox.\egroup
  }{2015}]{Chen:coco15}
Chen, X.; Fang, H.; Lin, T.; Vedantam, R.; Gupta, S.; Doll{\'{a}}r, P.; and
  Zitnick, C.~L.
\newblock 2015.
\newblock Microsoft {COCO} captions: Data collection and evaluation server.
\newblock {\em arXiv preprint arXiv:1504.00325}.

\bibitem[\protect\citeauthoryear{Chen \bgroup et al\mbox.\egroup
  }{2017}]{chen2017video}
Chen, S.; Chen, J.; Jin, Q.; and Hauptmann, A.
\newblock 2017.
\newblock Video captioning with guidance of multimodal latent topics.
\newblock In {\em ACM MM}.

\bibitem[\protect\citeauthoryear{Chen \bgroup et al\mbox.\egroup
  }{2018}]{Chen18:PickNet}
Chen, Y.; Wang, S.; Zhang, W.; and Huang, Q.
\newblock 2018.
\newblock Less is more: Picking informative frames for video captioning.
\newblock {\em CoRR} abs/1803.01457.

\bibitem[\protect\citeauthoryear{Dauphin \bgroup et al\mbox.\egroup
  }{2017}]{Dauphin:ICML17}
Dauphin, Y.~N.; Fan, A.; Auli, M.; and Grangier, D.
\newblock 2017.
\newblock Language modeling with gated convolutional networks.
\newblock In {\em ICML}.

\bibitem[\protect\citeauthoryear{Gehring \bgroup et al\mbox.\egroup
  }{2017}]{Gehring:ICML17}
Gehring, J.; Auli, M.; Grangier, D.; Yarats, D.; and Dauphin, Y.~N.
\newblock 2017.
\newblock Convolutional sequence to sequence learning.
\newblock In {\em ICML}.

\bibitem[\protect\citeauthoryear{Hao, Zhang, and Guan}{2018}]{Hao:aaai18}
Hao, W.; Zhang, Z.; and Guan, H.
\newblock 2018.
\newblock Integrating both visual and audio cues for enhanced video caption.
\newblock In {\em AAAI}.

\bibitem[\protect\citeauthoryear{He \bgroup et al\mbox.\egroup
  }{2016}]{He:CVPR16}
He, K.; Zhang, X.; Ren, S.; and Sun, J.
\newblock 2016.
\newblock Deep residual learning for image recognition.
\newblock In {\em CVPR}.

\bibitem[\protect\citeauthoryear{Kalchbrenner \bgroup et al\mbox.\egroup
  }{2016}]{Kalchbrenner:ByteNet16}
Kalchbrenner, N.; Espeholt, L.; Simonyan, K.; van~den Oord, A.; Graves, A.; and
  Kavukcuoglu, K.
\newblock 2016.
\newblock Neural machine translation in linear time.
\newblock {\em arXiv preprint arXiv:1610.10099}.

\bibitem[\protect\citeauthoryear{Karpathy \bgroup et al\mbox.\egroup
  }{2014}]{Karpathy:CVPR14}
Karpathy, A.; Toderici, G.; Shetty, S.; Leung, T.; Sukthankar, R.; and Li, F.
\newblock 2014.
\newblock Large-scale video classification with convolutional neural networks.
\newblock In {\em CVPR}.

\bibitem[\protect\citeauthoryear{Kingma and Ba}{2015}]{kingma2014adam}
Kingma, D., and Ba, J.
\newblock 2015.
\newblock Adam: A method for stochastic optimization.
\newblock In {\em ICLR}.

\bibitem[\protect\citeauthoryear{Li \bgroup et al\mbox.\egroup
  }{2018}]{li2018jointly}
Li, Y.; Yao, T.; Pan, Y.; Chao, H.; and Mei, T.
\newblock 2018.
\newblock Jointly localizing and describing events for dense video captioning.
\newblock In {\em CVPR}.

\bibitem[\protect\citeauthoryear{Li, Zhao, and Lu}{2017}]{Li17:MAMRNN}
Li, X.; Zhao, B.; and Lu, X.
\newblock 2017.
\newblock {MAM-RNN:} multi-level attention model based {RNN} for video
  captioning.
\newblock In {\em IJCAI}.

\bibitem[\protect\citeauthoryear{Meng \bgroup et al\mbox.\egroup
  }{2015}]{Meng:ACL15}
Meng, F.; Lu, Z.; Wang, M.; Li, H.; Jiang, W.; and Liu, Q.
\newblock 2015.
\newblock Encoding source language with convolutional neural network for
  machine translation.
\newblock In {\em ACL}.

\bibitem[\protect\citeauthoryear{Pan \bgroup et al\mbox.\egroup
  }{2016}]{Pan:CVPR16}
Pan, Y.; Mei, T.; Yao, T.; Li, H.; and Rui, Y.
\newblock 2016.
\newblock Jointly modeling embedding and translation to bridge video and
  language.
\newblock In {\em CVPR}.

\bibitem[\protect\citeauthoryear{Pan \bgroup et al\mbox.\egroup
  }{2017}]{Pan:CVPR17}
Pan, Y.; Yao, T.; Li, H.; and Mei, T.
\newblock 2017.
\newblock Video captioning with transferred semantic attributes.
\newblock In {\em CVPR}.

\bibitem[\protect\citeauthoryear{Papineni \bgroup et al\mbox.\egroup
  }{2002}]{Papineni:ACL02}
Papineni, K.; Roukos, S.; Ward, T.; and Zhu, W.
\newblock 2002.
\newblock Bleu: a method for automatic evaluation of machine translation.
\newblock In {\em ACL}.

\bibitem[\protect\citeauthoryear{Ramanishka \bgroup et al\mbox.\egroup
  }{2016}]{Ramanishka:ACMMM16}
Ramanishka, V.; Das, A.; Park, D.~H.; Venugopalan, S.; Hendricks, L.~A.;
  Rohrbach, M.; and Saenko, K.
\newblock 2016.
\newblock Multimodal video description.
\newblock In {\em ACM MM}.

\bibitem[\protect\citeauthoryear{Russakovsky \bgroup et al\mbox.\egroup
  }{2015}]{Russakovsky:IJCV15}
Russakovsky, O.; Deng, J.; Su, H.; Krause, J.; Satheesh, S.; Ma, S.; Huang, Z.;
  Karpathy, A.; Khosla, A.; Bernstein, M.~S.; Berg, A.~C.; and Li, F.
\newblock 2015.
\newblock Imagenet large scale visual recognition challenge.
\newblock {\em IJCV}.

\bibitem[\protect\citeauthoryear{Song \bgroup et al\mbox.\egroup
  }{2017}]{Song:IJCAI17}
Song, J.; Gao, L.; Guo, Z.; Liu, W.; Zhang, D.; and Shen, H.-T.
\newblock 2017.
\newblock Hierarchical {LSTM} with adjusted temporal attention for video
  captioning.
\newblock In {\em IJCAI}.

\bibitem[\protect\citeauthoryear{Tran \bgroup et al\mbox.\egroup
  }{2015}]{Tran:ICCV15}
Tran, D.; Bourdev, L.~D.; Fergus, R.; Torresani, L.; and Paluri, M.
\newblock 2015.
\newblock Learning spatiotemporal features with 3d convolutional networks.
\newblock In {\em ICCV}.

\bibitem[\protect\citeauthoryear{Vedantam, Zitnick, and
  Parikh}{2015}]{Vedantam:CVPR15}
Vedantam, R.; Zitnick, C.~L.; and Parikh, D.
\newblock 2015.
\newblock Cider: Consensus-based image description evaluation.
\newblock In {\em CVPR}.

\bibitem[\protect\citeauthoryear{Venugopalan \bgroup et al\mbox.\egroup
  }{2015a}]{Venugopalan:ICCV15}
Venugopalan, S.; Rohrbach, M.; Donahue, J.; Mooney, R.~J.; Darrell, T.; and
  Saenko, K.
\newblock 2015a.
\newblock Sequence to sequence - video to text.
\newblock In {\em ICCV}.

\bibitem[\protect\citeauthoryear{Venugopalan \bgroup et al\mbox.\egroup
  }{2015b}]{Venugopalan:NAACL15}
Venugopalan, S.; Xu, H.; Donahue, J.; Rohrbach, M.; Mooney, R.~J.; and Saenko,
  K.
\newblock 2015b.
\newblock Translating videos to natural language using deep recurrent neural
  networks.
\newblock In {\em NAACL}.

\bibitem[\protect\citeauthoryear{Wang \bgroup et al\mbox.\egroup
  }{2018}]{Wang18:M3}
Wang, J.; Wang, W.; Huang, Y.; Wang, L.; and Tan, T.
\newblock 2018.
\newblock M$^3$: Multimodal memory modelling for video captioning.
\newblock In {\em CVPR}.

\bibitem[\protect\citeauthoryear{Wu and Han}{2018}]{Wu18:MCNN}
Wu, A., and Han, Y.
\newblock 2018.
\newblock Multi-modal circulant fusion for video-to-language and backward.
\newblock In {\em IJCAI}.

\bibitem[\protect\citeauthoryear{Xu \bgroup et al\mbox.\egroup
  }{2016}]{Xu:CVPR16}
Xu, J.; Mei, T.; Yao, T.; and Rui, Y.
\newblock 2016.
\newblock {MSR-VTT:} {A} large video description dataset for bridging video and
  language.
\newblock In {\em CVPR}.

\bibitem[\protect\citeauthoryear{Xu \bgroup et al\mbox.\egroup
  }{2017}]{Xu:ACMMM17}
Xu, J.; Yao, T.; Zhang, Y.; and Mei, T.
\newblock 2017.
\newblock Learning multimodal attention {LSTM} networks for video captioning.
\newblock In {\em ACM MM}.

\bibitem[\protect\citeauthoryear{Yao \bgroup et al\mbox.\egroup
  }{2015}]{Yao:ICCV15}
Yao, L.; Torabi, A.; Cho, K.; Ballas, N.; Pal, C.~J.; Larochelle, H.; and
  Courville, A.~C.
\newblock 2015.
\newblock Describing videos by exploiting temporal structure.
\newblock In {\em ICCV}.

\bibitem[\protect\citeauthoryear{Yao \bgroup et al\mbox.\egroup
  }{2017a}]{yao2017novel}
Yao, T.; Pan, Y.; Li, Y.; and Mei, T.
\newblock 2017a.
\newblock Incorporating copying mechanism in image captioning for learning
  novel objects.
\newblock In {\em CVPR}.

\bibitem[\protect\citeauthoryear{Yao \bgroup et al\mbox.\egroup
  }{2017b}]{yao2017boosting}
Yao, T.; Pan, Y.; Li, Y.; Qiu, Z.; and Mei, T.
\newblock 2017b.
\newblock Boosting image captioning with attributes.
\newblock In {\em ICCV}.

\bibitem[\protect\citeauthoryear{Yao \bgroup et al\mbox.\egroup
  }{2018}]{yao2018exploring}
Yao, T.; Pan, Y.; Li, Y.; and Mei, T.
\newblock 2018.
\newblock Exploring visual relationship for image captioning.
\newblock In {\em ECCV}.

\bibitem[\protect\citeauthoryear{Yu \bgroup et al\mbox.\egroup
  }{2016}]{Yu:CVPR16}
Yu, H.; Wang, J.; Huang, Z.; Yang, Y.; and Xu, W.
\newblock 2016.
\newblock Video paragraph captioning using hierarchical recurrent neural
  networks.
\newblock In {\em CVPR}.

\bibitem[\protect\citeauthoryear{Zhao, Li, and Lu}{2018}]{Zhao18:tubefeatures}
Zhao, B.; Li, X.; and Lu, X.
\newblock 2018.
\newblock Video captioning with tube features.
\newblock In {\em IJCAI}.

\end{thebibliography}
\bibliographystyle{aaai}
\end{document}